\documentclass[11pt,a4paper]{article}
\usepackage{emnlp-ijcnlp-2019}
\usepackage{times}
\usepackage{latexsym}
\usepackage{url}
\usepackage{amssymb}
\usepackage{graphicx}
\usepackage{booktabs}
\usepackage{algorithm}
\usepackage{algorithmic}
\usepackage{multirow}
\usepackage{amsmath,bm}
\usepackage{url}
\usepackage{hyperref}
\usepackage{url}
\usepackage{helvet}
\usepackage{courier}
\usepackage{colortbl}
\usepackage{graphicx}
\usepackage{booktabs}       
\usepackage{amsfonts}       
\usepackage{nicefrac}       
\usepackage{microtype}      
\usepackage{verbatim}
\usepackage{amssymb}
\usepackage{pifont}

\usepackage{wrapfig}
\usepackage{lipsum}
\usepackage{url}

\usepackage{subfigure}
\newcommand*\samethanks[1][\value{footnote}]{\footnotemark[#1]}
\aclfinalcopy 

\title{A Closer Look at Data Bias in Neural Extractive Summarization Models}

\author{Ming Zhong\thanks{\hspace{1mm} These three authors contributed equally.},  Danqing Wang\samethanks, Pengfei Liu\samethanks, Xipeng Qiu\thanks{\ \  Corresponding author.} ,  Xuanjing Huang \\
  Shanghai Key Laboratory of Intelligent Information Processing, Fudan University \\
  School of Computer Science, Fudan University \\
  825 Zhangheng Road, Shanghai, China \\
  \texttt{\{mzhong18,dqwang18,pfliu14,xpqiu,xjhuang\}@fudan.edu.cn} }

\date{}

\begin{document}
\setlength{\abovedisplayskip}{2pt}
\setlength{\belowdisplayskip}{2pt}

\maketitle
\begin{abstract}
In this paper, we take stock of the current state of summarization datasets and explore how different factors of datasets influence the generalization behaviour of neural extractive summarization models. Specifically, we first propose several properties of datasets, which matter for the generalization of summarization models. Then we build the connection between priors residing in datasets and model designs, analyzing how different properties of datasets influence the choices of model structure design and training methods. Finally, by taking a typical dataset as an example, we rethink the process of the model design based on the experience of the above analysis. We demonstrate that when we have a deep understanding of the characteristics of datasets, a simple approach can bring significant improvements to the existing state-of-the-art model.
\end{abstract}

\section{Introduction}
\label{Introduction}

Neural network-based models have achieved great success on summarization tasks \cite{see2017get, celikyilmaz2018deep, jadhav2018extractive}.
Current studies on summarization either explore the possibility of optimization in terms of \textbf{networks' structures} \cite{zhou2018neural, chen2018fast, gehrmann2018bottom}, the improvement in terms of \textbf{training} \textbf{schemas} \cite{wangexploring2019,narayan2018ranking, wu2018learning, chen2018fast}, or the information fusion with large \textbf{pre-trained knowledge} \cite{peters2018deep,devlin2018bert,DBLP:journals/corr/abs-1903-10318,dong2019unified}.
More recently, \citet{zhong-etal-2019-searching} conducts  a comprehensive analysis  on why existing summarization systems  perform  so  well from above three aspects.
Despite their success, a relatively missing topic\footnote{Concurrent with our work, \cite{jungearlier2019} makes a similar analysis on  datasets biases and presents three factors which matter for the text summarization task.} is to
analyze and understand  the impact on the models' generalization ability from a dataset perspective.
With the emergence of more and more summarization datasets \cite{sandhaus2008new,nallapati2016abstractive,cohan2018discourse,grusky2018newsroom},
the time is ripe for us to bridge the gap between the insufficient understanding of the nature of \textit{datasets} themselves and the increasing improvement of the \textit{learning methods}.


In this paper, we  take a step towards addressing this challenge by taking neural extractive summarization models as an interpretable testbed, investigating how to quantify the characteristics of datasets. As a result, we could explain the behaviour of our models and design new ones. Specifically, we seek to answer two main questions:

\textbf{Q1:} In the summarization task, different datasets present diverse characteristics, so \textit{what is the bias introduced by these dataset choices and how does it influence the model's generalization ability?}
We explore two types of factors: \textit{constituent factors} and \textit{style factors}, and analyze how they affect the generalization of neural summarization models respectively.
These factors can help us diagnose the weakness of existing models.
\textbf{Q2}: \textit{How different  properties  of  datasets  influence  the choices of model structure design  and training schemas?}
We propose some measures and examine their abilities to explain how different model architectures, training schemas, and pre-training strategies react to various properties of datasets.

    \begin{table*}[htbp]
  \centering  \footnotesize
    \begin{tabular}{lllr}
    \toprule
    \textbf{Factors of Datasets} & \textbf{Measures} & \textbf{Model designs} &  \\
    \midrule
    \multicolumn{1}{l}{\multirow{2}[2]{*}{{Constituent}[\ref{sec:Constituentfactors}]}} & Positional coverage rate [\ref{pos}]   & Architecture designs [\ref{c-bias}] &  \\
          & Content  coverage  rate [\ref{con}]   & Pre-trained strategies [\ref{c-bias}] &  \\
    \midrule
    \multirow{2}[2]{*}{{Style [\ref{sec:stylefactors}]}} & Density [\ref{den}] & \multirow{2}[2]{*}{Training schemas [\ref{s-bias}]} &  \\
          & Compression [\ref{comp}] &       &  \\
    \bottomrule
    \end{tabular}%
    \caption{Organization structure of this paper: four measures presented in this paper and choices of model designs they have influence on. }
  \label{tab:addlabel}%
\end{table*}%

Our contributions can be summarized as follows:
\paragraph{Main Contributions}

1) For the summarization task itself, we diagnose the weakness of existing learning methods in terms of networks' structures, training schemas, and pre-trained knowledge. Some observations could instruct future researchers for a new state-of-the-art performance.
2)
We show that a comprehensive understanding of the dataset's properties guides us to design a more reasonable model.
We hope to encourage future research on how characteristics of datasets influence the behavior of neural networks.

We summarize our observations as follows:
1) Existing models under-utilize  the  nature  of  the  training  data. We demonstrate that a simple training method on CNN/DM (dividing training set based on domain) can achieve significant improvement.
2) BERT is not a panacea and will fail in some situation. The improvement brought by BERT is related to the \textit{style factor} defined in this paper.
3) It is difficult to handle the hard cases (defined by \textit{style factor}) via architecture design and pre-training knowledge under the extractive framework.
4) Based on the sufficient understanding of the nature of datasets, a more reasonable data partitioning (based on \textit{constituent factors}) method can be mined.

\section{Related Work}
We briefly outline connections and differences to following related lines of research.

\paragraph{Neural Extractive Summarization}
Recently, neural network-based models have achieved great success in extractive summarization. \cite{celikyilmaz2018deep, jadhav2018extractive, DBLP:journals/corr/abs-1903-10318}.
Existing works on text summarization can roughly fall into one of three classes: exploring \textit{networks' structures} with suitable bias \cite{cheng2016neural, nallapati2017summarunner, zhou2018neural}; introducing new \textit{training schemas} \cite{narayan2018ranking, wu2018learning, chen2018fast}  and incorporating large \textit{pre-trained knowledge} \cite{peters2018deep,devlin2018bert,DBLP:journals/corr/abs-1903-10318,dong2019unified}.
Instead of exploring the possibility for a new state-of-the-art along one of above three lines, in this paper, we aim to
bridge the gap
between the lack of understanding of the characteristics for the datasets
and the increasing development of above three learning methods.

Concurrent with our work, \cite{jungearlier2019} conducts a quite similar analysis on datasets biases and proposes three factors which matter for the text summarization task.
One major difference between these two works is that we additionally focus on how dataset biases influence the designs of models.

\paragraph{Understanding the Generalization Ability of Neural Networks}
While neural networks have shown superior generalization ability, yet it remains largely unexplained. Recently, some researchers begin to take a step towards understanding the generalization behaviour of neural networks from the perspective of network architectures or optimization procedure \cite{schmidt2018adversarially,baluja2017adversarial, zhang2016understanding, arpit2017closer}.
Different from these work, in this paper, we claim that interpreting the  generalization ability of neural networks is built on a good understanding of the characteristic of the data.

\section{Learning Methods and Datasets}

\subsection{Learning Methods}

Generally, given a dataset $\mathcal{D}$, different learning methods are trying to explain the data in diverse ways, which show different generalization behaviours.
Existing learning methods for extractive summarization systems vary in architectures designs, pre-trained strategies and training schemas.

\paragraph{Architecture Designs}
Architecturally speaking, most of existing extractive summarization systems consists of three major modules: \textbf{sentence encoder}, \textbf{document encoder} and \textbf{decoder}.

In this paper, our architectural choices vary with two types of document encoders: \texttt{LSTM\footnote{We use the implementation of \citet{he2017deep}.}} \cite{hochreiter1997long} and \texttt{Transformer} \cite{vaswani2017attention} while we keep the sentence encoder (convolutional neural networks) and decoder (sequence labeling) unchanged\footnote{Since they do not show significant influence on our explored experiments.}. The base model in all experiments refers to Transformer equipped with sequence labelling.

\paragraph{Pre-trained Strategies}

To explore how different pre-trained strategies influence the model, we take two types of pre-trained knowledge into consideration: we choose Word2vec \cite{mikolov2013efficient} as an investigated exemplar for non-contextualized word embeddings and adopt BERT as a contextualized word pre-trainer \cite{devlin2018bert}.

\paragraph{Training Schemas}
In general, we train a monolithic model to fit the dataset, but in particular, when the data itself has some special properties, we can introduce different training methods to fully exploit all the information contained in the data.

\begin{enumerate}
    \vspace{-2pt}
    \item \textbf{Multi-domain Learning}
    The basic idea of multi-domain learning in this paper is to introduce domain tag as a low-dimension vector which can augment learned representations.
    Domain-aware model will make it possible to learn domain-specific features.
    \vspace{-5pt}
    \item \textbf{Meta-learning}
    we also try to make models aware of different distribution by meta-learning based on \cite{wangexploring2019}. Specifically, for each iteration, we sample several domains as meta-train and the other as meta-test. The meta-test gradients will be combined with the meta-train gradients and finally update the model.
\end{enumerate}

\subsection{Datasets}
We explore four mainstream news articles summarization datasets (\texttt{CNN/DM}, \texttt{Newsroom}, \texttt{NYT50} and \texttt{DUC2002}) which are various in their publications. We also modify two large-scale scientific paper datasets (\texttt{arXiv} and \texttt{PubMed}) to investigate characteristics for different domains. Detailed statistics are illustrated in Table \ref{tab:dataset}.

\renewcommand\arraystretch{1.1}
\begin{table*}[htbp]\small
  \centering
    \setlength{\tabcolsep}{1.5mm}{
    \begin{tabular}{lrrrrrrrrrrr}
    \toprule
          & \multicolumn{3}{c}{\textbf{Statistics}} & \multicolumn{2}{c}{\textbf{Measures}} & \multicolumn{3}{c}{\textbf{Lead-$k$}} & \multicolumn{3}{c}{\textbf{Ext-Oralce}} \\
    \midrule
          & \multicolumn{1}{c}{\textbf{Train}} & \multicolumn{1}{c}{\textbf{Valid}} & \multicolumn{1}{c}{\textbf{Test}} & \multicolumn{1}{c}{\textbf{Density}} & \multicolumn{1}{c}{\textbf{Compres.}} & \multicolumn{1}{c}{\textbf{R-1}} & \multicolumn{1}{c}{\textbf{R-2}} & \multicolumn{1}{c}{\textbf{R-L}} & \multicolumn{1}{c}{\textbf{R-1}} & \multicolumn{1}{c}{\textbf{R-2}} & \multicolumn{1}{c}{\textbf{R-L}} \\
    \midrule
    CNN/DM (3) & \cellcolor[rgb]{ .851,  .851,  .851}  287,227  & \cellcolor[rgb]{ .851,  .851,  .851}      13,368  & \cellcolor[rgb]{ .851,  .851,  .851}     11,490  & 3.70  & 13.76 & \cellcolor[rgb]{ .851,  .851,  .851}40.24 & \cellcolor[rgb]{ .851,  .851,  .851}17.53 & \cellcolor[rgb]{ .851,  .851,  .851}36.29 & 56.55 & 33.40 & 53.03 \\
    arXiv (6) & \cellcolor[rgb]{ .851,  .851,  .851}  187,324  & \cellcolor[rgb]{ .851,  .851,  .851}        6,218  & \cellcolor[rgb]{ .851,  .851,  .851}       6,217  & 2.19  & 5.59  & \cellcolor[rgb]{ .851,  .851,  .851}35.37 & \cellcolor[rgb]{ .851,  .851,  .851}9.25 & \cellcolor[rgb]{ .851,  .851,  .851}30.93 & 52.44 & 22.72 & 46.15 \\
    PubMed (5) & \cellcolor[rgb]{ .851,  .851,  .851}    87,897  & \cellcolor[rgb]{ .851,  .851,  .851}        4,946  & \cellcolor[rgb]{ .851,  .851,  .851}       5,031  & 2.04  & 2.28  & \cellcolor[rgb]{ .851,  .851,  .851}36.09 & \cellcolor[rgb]{ .851,  .851,  .851}11.49 & \cellcolor[rgb]{ .851,  .851,  .851}32.13 & 46.19 & 19.91 & 40.83 \\
    DUC2002 (6) & \cellcolor[rgb]{ .851,  .851,  .851} -  & \cellcolor[rgb]{ .851,  .851,  .851} -  & \cellcolor[rgb]{ .851,  .851,  .851}        567  & 4.43  & 5.52  & \cellcolor[rgb]{ .851,  .851,  .851}47.65 & \cellcolor[rgb]{ .851,  .851,  .851}23.19 & \cellcolor[rgb]{ .851,  .851,  .851}43.92 & 62.15 & 37.30 & 58.33 \\
    NYT50 (4) & \cellcolor[rgb]{ .851,  .851,  .851}    96,826  & \cellcolor[rgb]{ .851,  .851,  .851}        4,000  & \cellcolor[rgb]{ .851,  .851,  .851}       3,452  & 4.64  & 15.33 & \cellcolor[rgb]{ .851,  .851,  .851}38.54 & \cellcolor[rgb]{ .851,  .851,  .851}19.90 & \cellcolor[rgb]{ .851,  .851,  .851}35.27 & 63.97 & 43.51 & 60.70 \\
    Newsroom (2) & \cellcolor[rgb]{ .851,  .851,  .851}  995,041  & \cellcolor[rgb]{ .851,  .851,  .851}    108,837  & \cellcolor[rgb]{ .851,  .851,  .851}   108,862  & 8.60  & 35.07 & \cellcolor[rgb]{ .851,  .851,  .851}34.19 & \cellcolor[rgb]{ .851,  .851,  .851}23.42 & \cellcolor[rgb]{ .851,  .851,  .851}31.41 & 54.97 & 41.28 & 51.62 \\
    \bottomrule
    \end{tabular}}%
    \caption{Detailed statistics of six datasets. Density and Compression are \textit{style factors} in Section \ref{sec:stylefactors}. Lead-$k$ indicates ROUGE score of the first $k$ sentences in the document and Ext-Oracle indicates ROUGE score of sentences in the ground truth, they represent the lower and upper bound of extractive models respectively. The figure in parentheses after the datasets denotes the number of sentences extracted in Lead-$k$, which is close to the average number of Ext-Oracle labels.}
  \label{tab:dataset}%
\end{table*}%

\section{Quantifying Characteristics of Text Summarization Datasets}
\label{4}

In this paper, we present four measures to quantify the characteristics of summarization datasets, which can be abstracted into two types:
\textit{constituent factor} and \textit{style factors}.

\subsection{Constituent Factors}
\label{sec:Constituentfactors}
\paragraph{Motivation}
When the neural summarization model determines whether a sentence should be extracted, the representation of the sentence consists of two components: position representation\footnote{The position representation is obtained from the model structure in LSTM and by positional embedding in Transformer.}, which indicates the position of the sentence in the document; content representation, which contains the semantic information of the sentence.

Therefore, we define the position and content information of the sentence as constituent factors, aiming to explore how the selected sentences in the test set relate to the training set in terms of position  and content information.


\subsubsection{Positional Information}  \label{pos}
\paragraph{Positional Value (P-Value)}
Given a document $D = {s_1, \cdots, s_n}$, for each sentence $s_i$ with label $y_i = 1$, we introduce the notion of positional value $p_i \in {1,\cdots,K}$, whose value is the output of the mapping function $p_i = \mathrm{f}(i)$.

\vspace{-5pt}
\paragraph{Positional Coverage Rate (PCR)}
Taking positional value $p$ as a discrete random variable, we can define the discrete probability distribution of $p$ over a dataset $\mathcal{D}$,

\begin{align}
    P(p=u) = \frac{N_u}{N_{sent}}
\end{align}
where $N_u$ denotes the number of sentence with $p = u$ and $N_{sent}$ represents the number of sentences with $y_i = 1$ in dataset $\mathcal{D}$.

Based on above definition, for any two datasets $\mathcal{D}^A$ and $\mathcal{D}^B$, we could  quantify the proximity of their positional value distribution
\begin{align}
\eta_p(\mathcal{D}^A,\mathcal{D}^B) = -log(\mathrm{KL}(P^{A}||P^{B}))
\end{align}

where $\mathrm{KL}(\cdot)$ denotes $\mathrm{KL}$-divergence function. $P^{A}$ and $P^{B}$ represent two position value distribution over two datasets. The datasets with similar positional value distribution usually have large PCR $\eta_p$.

\subsubsection{Content Information}  \label{con}
\paragraph{Content Value (C-Value)}
Given a dataset $\mathcal{D}$, we want to find the patterns that appear most frequently in the ground truth\footnote{Ground truth is extracted by the greedy algorithm in \citet{nallapati2017summarunner}} of $\mathcal{D}$ and score them. For each sentence in gound truth, we remove the stop words and punctuation, replace all numbers with ``0", and perform lemmatization on each token. After the pre-processing, we treat $n$-gram ($n>1$) as the pattern in $\mathcal{D}$ and calculate the score $\varphi(pt_i, \mathcal{D})$ for each pattern as follows:
\begin{align} \small
    {\varphi(pt_i, \mathcal{D})} = \frac{N_{pt_i}}{\sum\limits_{pt_j \in \mathcal{D} } N_{pt_j}}
\end{align}

where $N_{pt_i}$ denotes the number of $i$-th pattern.

\vspace{-5pt}
\paragraph{Content Coverage Rate (CCR)}
We introduce the notion of $\eta_c$ to measure the degree of contents' overlap between training and test set in which the sentences with ground truth labels reside in.
\begin{align} \small
    \mathrm{Sim(i, j)} = \mathrm{\varphi(pt_i, \phi_{tr})} * \mathrm{\varphi(pt_j, \phi_{te})}
\end{align}
\begin{align} \small
    \eta_c(\mathcal{D}_{tr},\mathcal{D}_{te}) =
    \sum_{pt_i \in \phi_{tr} } \sum_{pt_j \in \phi_{te} } \mathrm{Sim(i, j)}
\end{align}
where $\phi$ denotes the set\footnote{We choose 100 bigrams and trigrams as the set.} of patterns which is helpful to pick out ground truth sentences. $\mathrm{Sim(\cdot)}$ measures the similarity of two patterns, $\mathcal{D}_{tr}$ and $\mathcal{D}_{te}$ represent the training set and test set of $\mathcal{D}$ respectively.

\subsection{Style Factors}
\label{sec:stylefactors}

\paragraph{Motivation}
Different from constituent factors, style factors influence the generalization ability of summarization models by adjusting the learning difficulty of samples' features.

For this type of factor, we did not propose a new measure, but adopt the indicators \textsc{density}, \textsc{compression} proposed by \cite{grusky2018newsroom}\footnote{\textsc{density} and \textsc{compression} was originally used to describe the diversity between datasets in the construction of new datasets.}
We claim that the contribution here is to focus on the understanding of these metrics and explore the reasons why they affect the performance of summarization models, which is missing from previous work. More importantly, only when we understand how these metrics affect the performance of the models can we use them to explain some of the differences in model generalization.

\subsubsection{Density}  \label{den}
Density is used to qualitatively measure the degree to which a summary is derivative of a document \cite{grusky2018newsroom}.
Specifically, given a document $D$ and its corresponding summary $S$, Density(D,S) measures the percentage of words in the summary that are from document.
\begin{align}
    \mathrm{Density}(D, S) = \frac{1}{|S|} \sum_{f\in \mathcal{F}(D, S)}|f|^2
\end{align}
where $|\cdot|$ denotes the number of words. ${\mathcal{F}(D,S)}$ is a set of extractive fragments, which characterize the the longest shared token sequence.

\subsubsection{Compression} \label{comp}
Compression is used to characterize the word ratio between the document and summary \cite{grusky2018newsroom}.
\begin{align}
    \mathrm{Compression}(d,s) = |D|\,/\,|S|
\end{align}

\section{Investigating Influence of Proposed Factors on Summarization Models}

\subsection{Constituent Factors}
\subsubsection{Exp-I: Breaking Down the Test set}
\label{4.1.3}

For the P-Value, the threshold set can be denoted as $\{t_0=0,t_1, \cdots ,t_K=\infty\}$. We calculate $Pos(i)$ for each sentence $s_i$:
\begin{align}
Pos(i)=
\left\{
    \begin{array}{lr}
     i &  0 \leq i < t_1\text{ or }i \geq t_{K-1}\\
     \frac{i}{n} \cdot t_{K-1} &  others
     \end{array}
\right.
\end{align}
and define $p_i=k$ if $t_{k-1} \leq Pos(i) < t_{k}$. The $Pos(i)$ considers both absolute and relevant position of the sentence in the document. In the experiment, we make $K=5$ and choose $\{0, 3, 7, 15, 35, \infty\}$ for the threshold set.

For the C-Value, we calculate the score for each sentence based on the pattern score from training set.
\begin{align} \small
    \mathrm{\varphi(s_i, \mathcal{D})} = \sum_{pt_j \in s_i} \mathrm{\varphi(pt_j, \mathcal{D}_{tr}}) 
\end{align}

where $s_i$ denotes sentence in the ground truth of test set. The score indicates the degree of overlap between the sentence and important patterns of the training set. We then sort all the sentences in ascending order by score and divide them into five intervals with the same number of sentences.

\begin{figure}[t]
  \centering
  \subfigure[P-Value]{
    \label{fig:pv}
    \includegraphics[width=0.45\linewidth]{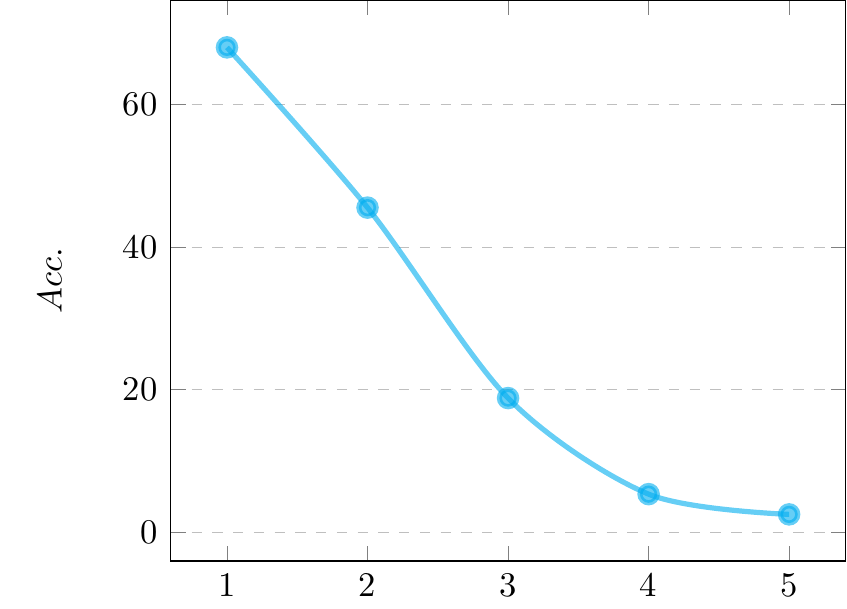}
  }
  \subfigure[C-Value]{
    \label{fig:cv}
    \includegraphics[width=0.4\linewidth]{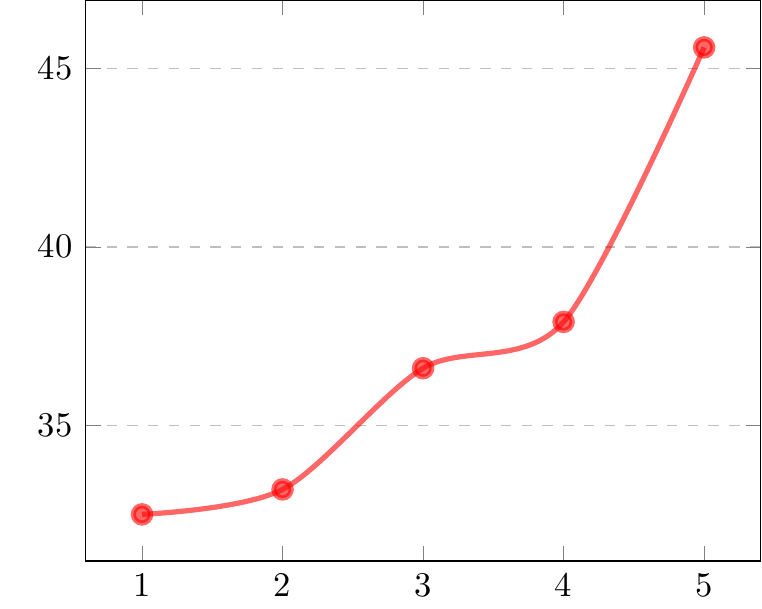}
  }
 \caption{The accuracy on CNN/DM dataset, test set is broken down based on P-Value and C-Value. }
 \label{fig:pv&cv}
\end{figure}

As shown in Figure \ref{fig:pv&cv}, when the sentence is in the front of the document or contains more salient patterns, the accuracy of the model to extract sentences is higher. The phenomenon means that our proposed P-Value and C-Value reflect position distribution and content information of a specific dataset to a certain extent, and the model does learn \textit{constituent factors} and uses them to determine whether a sentence is selected.

\subsubsection{Exp-II: Cross-dataset Generalization}
\label{p-cf}
From the above experiments, we can see that P-Value and C-Value are sufficient to characterize some attributes in a specific dataset, but beyond that, we seek to understand the differences between mainstream datasets through PCR and CCR.

\renewcommand\arraystretch{1.0}
\begin{table}[htbp]\footnotesize\setlength{\tabcolsep}{2.3pt}
  \centering
    \begin{tabular}{llccccc}
    \toprule
      & Dataset & CNNDM & arXiv & Pubmed & NYT50 & Newsr. \\
    \midrule
    \multirow{5}[2]{*}{PCR} & CNNDM & \cellcolor[rgb]{ .851,  .851,  .851}1.41 & 0.56  & 0.38  & 0.75  & 0.70 \\
          & arXiv & 0.51  & \cellcolor[rgb]{ .851,  .851,  .851}2.38 & 0.68  & 0.15  & 0.28 \\
          & Pubmed & 0.38  & 0.66  & \cellcolor[rgb]{ .851,  .851,  .851}3.79 & 0.08  & 0.41 \\
          & NYT50 & 1.27  & 0.22  & 0.23  & \cellcolor[rgb]{ .851,  .851,  .851}1.46 & 1.28 \\
          & Newsr. & 1.02  & 0.30  & 0.40  & 0.79  & \cellcolor[rgb]{ .851,  .851,  .851}4.57 \\
    \midrule
    \multirow{5}[2]{*}{CCR} & CNNDM & \cellcolor[rgb]{ .851,  .851,  .851}3.69 & 0.07  & 0.89  & 1.32  & 1.56 \\
          & arXiv & 0.05  & \cellcolor[rgb]{ .851,  .851,  .851}10.04 & 0.47  & 0.03  & 0.16 \\
          & Pubmed & 0.72  & 0.62  & \cellcolor[rgb]{ .851,  .851,  .851}11.03 & 0.51  & 2.03 \\
          & NYT50 & 1.34  & 0.07  & 0.75  & \cellcolor[rgb]{ .851,  .851,  .851}3.13 & 2.12 \\
          & Newsr. & 1.27  & 0.21  & 2.09  & 1.41  & \cellcolor[rgb]{ .851,  .851,  .851}4.21 \\
    \midrule
    \multirow{5}[1]{*}{R-2} & CNNDM & \cellcolor[rgb]{ .851,  .851,  .851}18.71 & 9.55  & 11.60 & 21.72 & 15.89 \\
          & arXiv & 11.46 & \cellcolor[rgb]{ .851,  .851,  .851}16.91 & 16.21 & 15.10 & 15.93 \\
          & PubMed & 9.68  & 15.56 & \cellcolor[rgb]{ .851,  .851,  .851}16.46 & 10.39 & 12.16 \\
          & NYT50 & 17.01 & 9.62  & 11.98 & \cellcolor[rgb]{ .851,  .851,  .851}25.39 & 20.52 \\
          & Newsr. & 17.38 & 9.42  & 12.23 & 20.21 & \cellcolor[rgb]{ .851,  .851,  .851}24.59 \\
    \bottomrule
    \end{tabular}%
  \caption{Results of cross-dataset PCR($\eta^p$), CCR($\eta^c$) and ROUGE-2 score. Each cell $\eta^p_{ij}$ and $\eta^c_{ij}$ denotes the coverage rate between training dataset (rows) and test dataset (columns). Each cell R-2$_{ij}$ denotes model performance in cross-dataset setting.}
  \label{tab:PCR & CCR & R2}
\end{table}%

We calculate PCR/CCR score and measure the performance of the base model by ROUGE-2 score on five datasets. We can see from Table \ref{tab:PCR & CCR & R2} that the training and test set of the same dataset always have the highest PCR/CCR score, which indicates the distribution between them is the closest based on \textit{consitituent factors}. Furthermore, model performance is also in accord with this trend. Consistency presented by the experiment, on the one hand, illustrates that there are significant shifts between different datasets, which results in performance differences of the model in cross-dataset setting, on the other hand, it reflects that position distribution and content information are the key factors of such dataset-shift.

After verifying the validity of PCR and CCR, we utilize them to estimate the distance between the real distribution of datasets. For instance, news articles datasets (CNN/DM, NTY50 and Newsroom) and scientific paper datasets (arXiv and PubMed) both have lower scores in terms of two metrics, that is to say, there is a larger shift between them, which is also in line with our knowledge. Based on the estimation, we can understand more deeply the impact of different datasets on the generalization ability of various neural extractive summarization models.

\subsection{Style Factors} \label{p-sf}
We integrate training set, validation set and test set as a whole set and divide it into three parts according to the density or compression of each article and name them ``low'', ``medium'' and ``high''. For example, articles in ``density, high'' represents these articles have a higher density in the entire dataset. Based on above operation, we break down the test set and attempt to analyze how style factors influence the model performance.

\renewcommand\arraystretch{1.1}
\begin{table}[htbp]
  \centering  \footnotesize
    \begin{tabular}{llccc}
    \toprule
          & \textbf{Metrics} & \textbf{Low} & \textbf{Medium} & \textbf{High} \\
    \midrule
    \multirow{3}[2]{*}{\textsc{Density}}
          & R-1    & 35.07 & 41.63 & \textbf{46.48} \\
          & R-2    & 11.22 & 17.87 & \textbf{26.39} \\
          & R-L    & 31.19 & 37.69 & \textbf{43.21} \\
          & F1    & 33.81 & 38.02 & \textbf{38.38} \\
    \midrule
    \multirow{3}[2]{*}{\textsc{Compres.}}
          & R-1    & \textbf{44.95} & 41.32 & 36.41 \\
          & R-2    & \textbf{21.74} & 18.61 & 15.10 \\
          & R-L    & \textbf{41.09} & 37.61 & 32.98 \\
          & F1    & \textbf{39.82} & 37.01 & 32.23 \\
    \bottomrule
    \end{tabular}%
    \caption{The performance of our base model on CNN/DM dataset, test set is broken down based on \textsc{Density} and \textsc{Compression}}
  \label{tab:breakdown performace}%
\end{table}%

\paragraph{Exploration of Density} Density represents the overlap between the summary and the original text, so the samples with high density are more friendly to extractive models. Consequently, it is easy for us to understand the higher the density, the higher the ROUGE score in Table \ref{tab:breakdown performace}. However, the {$\rm F_1$} value of prediction is also positively correlated with the density, which means that density is closely related to the learning difficulty.

\renewcommand\arraystretch{1.1}
\begin{table}[htbp]\small\setlength{\tabcolsep}{7pt}
  \centering

    \begin{tabular}{clllll}
    \toprule
          &       & \textbf{1} & \textbf{2} & \textbf{3} & \textbf{Total}  \\
    \midrule
    \multirow{2}[2]{*}{\textbf{Low}} & $\psi$ & 0.18  & 0.15  & 0.13  & 0.46 \\
          & Pct & 8.5\%  & 7.0\%  & 6.0\%  & 21.5\% \\
    \midrule
    \multirow{2}[2]{*}{\textbf{Medium}} & $\psi$ & 0.24  & 0.20  & 0.17  & 0.61 \\
          & Pct & 9.0\%  & 7.6\%  & 6.4\%   & 23.0\% \\
    \midrule
    \multirow{2}[2]{*}{\textbf{High}} & $\psi$ & 0.32  & 0.27  & 0.22  & 0.81  \\
          & Pct & 10.8\%  & 9.1\%  & 7.5\%   & 27.4\% \\
    \bottomrule
    \end{tabular}%
    \caption{Experiment about \textsc{density}, Pct denotes the percentage of $\psi(s_i, S)$ to $\sum_{s_i \in D}\psi(s_i, S)$. The first three sentences contain more salient information in samples with higher density.}
  \label{tab:LCS}%
\end{table}%

In order to comprehend this correlation, we conduct the following experiment. Given an article and summary pair, we assign a score $\psi(s_i, S)$ to each sentence in article to indicate how much sailent information is contained in the sentence.
\begin{align} \small
    {\psi(s_i, S)} = {\mathrm{LCS(s_i, S)}} \,/\, {|s_i|}
\end{align}
where $\mathrm{LCS(s_i, S)}$ denotes the longest common subsequence length (not counting stop words and punctuation) of the sentence and summary. We calculate the percentage of $\psi(s_i, S)$ to $\sum_{s_i \in D}\psi(s_i, S)$ and present the results of the three highest-scoring sentences in Table \ref{tab:LCS}. Obviously, in samples with high density, the salient information is more concentrated in a few sentences, making it easier for the model to extract correct sentences.

Therefore, for dataset with high density, we can try to introduce external knowledge into the model, which helps the model better understand the semantic information, and thus easier to capture sentences with salient patterns. In addition, models with external knowledge should have better generalization ablity when transferred to high-density dataset. These inferences will be verified in Section \ref{s-bias} and \ref{5.2.1}.

\vspace{-5pt}
\paragraph{Exploration of Compression} Documents with high compression tend to have fewer sentences because summaries usually have a similar length in the same dataset. So the results of compression in Table \ref{tab:breakdown performace} are in line with our expectations, how the model represents long documents to get good performance in text summarization task remains a challenge \cite{celikyilmaz2018deep}.

\renewcommand\arraystretch{1.0}
\begin{table*}[!htbp]\small\setlength{\tabcolsep}{10pt}
  \centerline{
    \begin{tabular}{lccc|ccc|c}
    \toprule
          \multirow{2}{*}{\textbf{Models}} & \multicolumn{3}{c}{\textbf{\textsc{Density}}}                                          & \multicolumn{3}{c}{\textbf{\textsc{Compression}}} & \multirow{2}{*}{\textbf{All}} \\
          \cmidrule{2-7}
            & \multicolumn{1}{c}{\textbf{Low}} & \multicolumn{1}{c}{\textbf{Medium}} & \multicolumn{1}{c}{\textbf{High}} & \multicolumn{1}{c}{\textbf{Low}} & \multicolumn{1}{c}{\textbf{Medium}} & \multicolumn{1}{c}{\textbf{High}} &  \\
    \midrule
            LSTM &  11.17 & 17.75 & 25.84 & 21.66 & 18.25 & 14.73 & 18.58 \\
            \quad -$\;$Word2Vec & $\;\,$0.08 $\downarrow$ & $\;\,$0.36 $\downarrow$ & $\;\,$0.57 $\downarrow$ & $\;\,$0.51 $\downarrow$ & $\;\,$0.32 $\downarrow$& $\;\,$0.18 $\downarrow$ & $\;\,$0.35 $\downarrow$ \\
            \quad + BERT & $\;\,$0.44 $\uparrow$ & $\;\,$0.80 $\uparrow$ & $\;\,$1.46 $\uparrow$ & $\;\,$1.19 $\uparrow$ & $\;\,$0.80 $\uparrow$ & $\;\,$0.68 $\uparrow$ & $\;\,$0.92 $\uparrow$ \\
    \midrule
            Transformer     & 11.15     & 17.84                    & 26.17 & 22.00 & 18.34 & 14.65 & 18.71 \\
            \quad -$\;    $Word2Vec   & $\;\, $0.10 $\downarrow$ & $\;\,$0.30 $\downarrow$ & $\;\,$0.54 $\downarrow$ & $\;\,$0.48 $\downarrow$ & $\;\,$0.12 $\downarrow$ & $\;\,$0.30 $\downarrow$ & $\;\,$0.31 $\downarrow$ \\
            \quad + BERT & $\;\,$0.52 $\uparrow$ & $\;\,$0.77 $\uparrow$ & $\;\,$1.28 $\uparrow$ & $\;\,$0.98 $\uparrow$ & $\;\,$0.86 $\uparrow$ & $\;\,$0.79 $\uparrow$ & $\;\,$0.88 $\uparrow$ \\
    \bottomrule
    \end{tabular}}
    \caption{Performance of models equipped with different types of knowledge on CNN/DM dataset. BERT here removes the gradient as a way of introducing external knowledge.}
    \label{tab:style-experiment}
\end{table*}

Unlike the exploration of density, we attempt to understand how the model extracts sentences when faced with different compression samples. We utilize an attribution technique called Integrated Gradients (IG) \cite{sundararajan2017axiomatic} to separate the position and content information of each sentence. The setting of \textit{input x} and \textit{baseline x'} in this paper is close to \citet{mudrakarta2018did}\footnote{Using empty documents (a sequence of word embeddings corresponding
to padding value) as \textit{baseline x'}.}, but it is worth noting that our base model adds positional embedding to each sentence, so \textit{input x} and \textit{baseline x'} both have positional information.

\begin{figure}[t]
  \centering
  \subfigure[Position attribution]{
    \label{fig:c_attr}
    \includegraphics[width=0.45\linewidth]{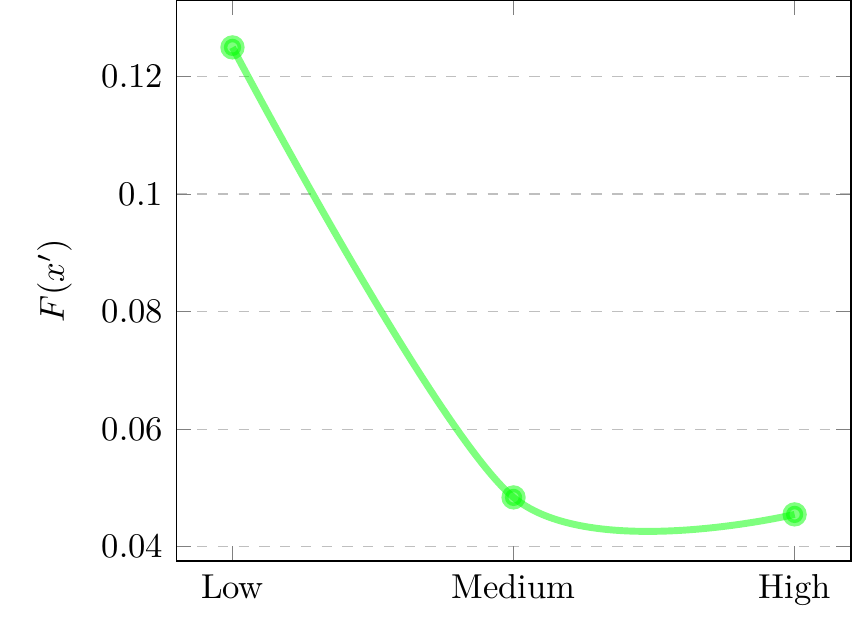}
  }
  \subfigure[Content attribution]{
    \label{fig:p_attr}
    \includegraphics[width=0.45\linewidth]{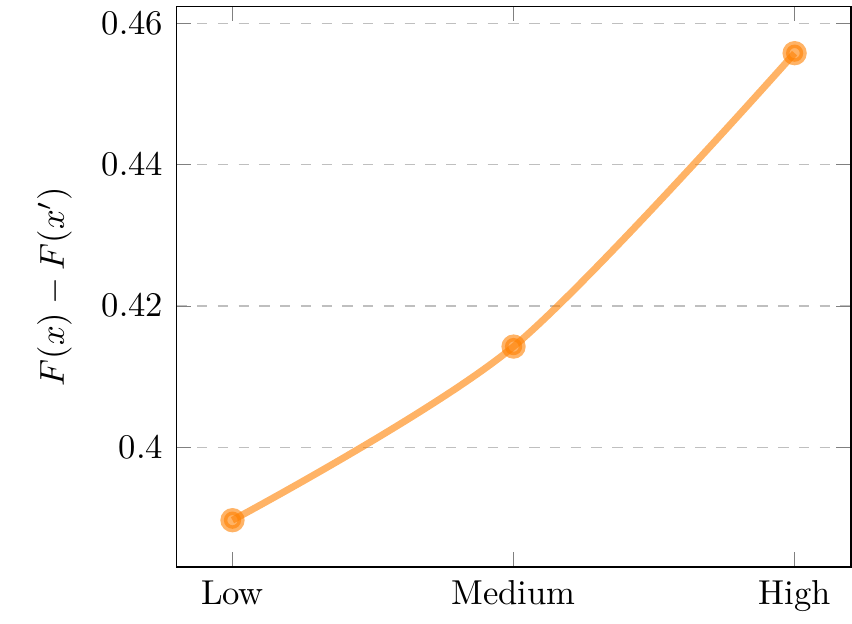}
  }
 \caption{The position and content attribution on CNN/DM, the test set is broken down based on \textsc{compression}. }
 \label{fig:attribution}
\end{figure}

We tend to think that $\mathrm{F(x')}$ denotes the attribution of positional information, and $\mathrm{F(x)}$ - $\mathrm{F(x')}$ denotes the attribution of content information when model makes decisions, where $\mathrm{F}: \mathbb{R}^{n} \rightarrow[0,1]$ represents a deep network. Figure \ref{fig:attribution} illustrates that as compression increases, the help provided by positional information is gradually reduced and content information becomes more important to the model. In other words, the model can perceive the compression and decide whether to pay more attention to positional information or important patterns, this observation is helpful for us to design models or study their generalization ability in Section \ref{5.2.1}.


\section{Bridge the Gap  between  Dataset Bias and Model Design Prior}
\label{sec:Experiment}

In this section we investigate how different properties of datasets influence the choices of model structures, pre-trained strategies, and training schemas.

\paragraph{Idea of Experiment Design}

Through the above analysis in Section \ref{4}, the \textit{constituent factors} reflect the relationship between diverse data distributions and \textit{style factors} directly affect the learning difficulty of samples' features.
Based on the different attributes of the above two types of factors, we designed the following investigation accordingly:
for the \textit{style factor}, we not only investigate the influence of different model architectures and pre-trained strategies on it, but
utilize it to explain the generalization behaviour of the models.
For the \textit{constituent factors}, we discuss their effects on different training strategies, such as multi-domain learning and meta-learning, because these learning modes are all about how to better model various types of distributions.

\vspace{-5pt}
\subsection{\textit{Style Factors}  Bias}
\label{s-bias}
In this section, we study \textbf{whether the samples with different learning difficulties described by the \textit{style factors} can be well handled through the improvement of structure or the introducing of pre-training knowledge or we need to extend our model in other ways}.

\begin{table*}[htbp]\small\setlength{\tabcolsep}{5pt}
  \centerline{
    \begin{tabular}{lccc|ccc|ccc|ccc}
    \toprule
    \multirow{2}{*}{\textbf{Dataset}}
     & \multicolumn{3}{c}{\textbf{Basic}} & \multicolumn{3}{c}{\textbf{Tag}} & \multicolumn{3}{c}{\textbf{Meta}} & \multicolumn{3}{c}{\textbf{Bert}} \\
    \cmidrule{2-13} & \multicolumn{1}{c}{\textbf{R-1}} & \multicolumn{1}{c}{\textbf{R-2}} & \multicolumn{1}{c}{\textbf{R-L}} & \multicolumn{1}{c}{\textbf{R-1}} & \multicolumn{1}{c}{\textbf{R-2}} & \multicolumn{1}{c}{\textbf{R-L}} & \multicolumn{1}{c}{\textbf{R-1}} & \multicolumn{1}{c}{\textbf{R-2}} & \multicolumn{1}{c}{\textbf{R-L}} & \multicolumn{1}{c}{\textbf{R-1}} & \multicolumn{1}{c}{\textbf{R-2}} & \multicolumn{1}{c}{\textbf{R-L}} \\
    \midrule
    CNNDM & 41.31 & 18.71 & 37.62 & \textcolor{red}{41.37} & \textcolor{red}{18.89} & \textcolor{red}{37.70} & 41.26 & 18.77 & 37.60 & \textbf{42.27} & \textbf{19.72} & \textbf{38.62} \\
    \midrule
    Arxiv & 36.29 & 9.55  & 32.01 & \textcolor{red}{\textbf{37.12}} & \textcolor{red}{10.09} & \textcolor{red}{\textbf{32.73}} & 36.25 & 9.58  & 31.95 & 37.00 & \textbf{10.22} & 32.59 \\
    Pubmed & 36.13 & 11.60 & 31.91 & \textcolor{red}{36.75} & \textcolor{red}{12.02} & \textcolor{red}{32.47} & 36.07 & 11.65 & 31.89 & \textbf{36.93} & \textbf{12.15} & \textbf{32.71} \\
    DUC2002 & 49.21 & 24.08 & 45.37 & 49.25 & \textcolor{red}{24.28} & 45.38 & \textcolor{red}{49.34} & 24.22 & \textcolor{red}{45.50} & \textbf{49.77} & \textbf{24.67} & \textbf{45.87} \\
    NYT50 & 41.17 & \textcolor{red}{21.72} & 37.85 & 40.84 & 21.06 & 37.45 & \textcolor{red}{41.23} & 21.56 & \textcolor{red}{37.89} & \textbf{43.80} & \textbf{24.09} & \textbf{40.46} \\
    Newsroom & 28.08 & \textcolor{red}{15.89} & \textcolor{red}{25.08} & 26.37 & 13.78 & 23.21 & \textcolor{red}{28.10} & 15.85 & 25.06 & \textbf{29.30} & \textbf{16.73} & \textbf{26.16} \\
    \bottomrule
    \end{tabular}}%
  \caption{Results of four models under two types of evaluation settings: \textsc{in-dataset}, and \textsc{cross-dataset}. Bold indicates the best performance of all models, red indicated the best performance other than BERT.}
    \label{tab:4models}%
\end{table*}%

Table \ref{tab:style-experiment} shows the breakdown performance on CNN/DM based on \textsc{density} and \textsc{compression}. And we can observe that:
1) An obvious trend is that LSTM performs better than Transformer with increasing difficulty in sample learning (low density and high compression).
For instance, LSTM performs worse than Transformer on the subset with high density, while surpasses Transformer when the density of testing examples becomes lower.
2) Generally, the introducing of pre-training word vectors can improve the overall results of the models. However, we found that increasing the learning difficulty of samples would weaken the benefits brought by pre-trained embeddings.
3) The prospects for further gains for these hard cases described by \textit{style factor} from novel architecture design and knowledge pre-training seem quite limited, suggesting that perhaps we should explore other ways, such as generating summaries instead of extracting.

\subsection{\textit{Constituent Factors} Bias} \label{c-bias}
We design our experiment towards the answer to two main questions as follows.

\subsubsection{Exp-I: How do dataset properties influence the choices of training schemas?}
\label{5.2.1}
When our training set itself contains multiple domains grouped by the \textit{constituent factors},
{how can we make full use of the dataset’s characteristic and find the most suitable training schemas?}
For example,  CNN/DailyMail, as one of the most popular datasets, consists of two sub-datasets.
For this question, dataset-shift discussed in Section \ref{p-cf} and the learning diffuculties of the dataset should be taken into consideration.

\vspace{-5pt}
\paragraph{Choices of Training Schemas:} We compare four training schemas: joint training, multi-domain learning\footnote{We view CNN and DailyMail in CNN/DM as two different domains.} with explicit information (tag embedding), implicit information (BERT) and meta-learning.

\vspace{-5pt}
\paragraph{Evaluation Setting:} In order to more comprehensively reflect the generalization ability of different models, we conducted zero-shot transfer evaluation. Specifically, each of our models is trained on \texttt{CNN/DM} while evaluated both on \texttt{CNN/DM} (\textsc{in-dataset}) and other datasets (\textsc{cross-dataset}).

Table \ref{tab:4models} shows the results of four models under two types of evaluation settings: \textsc{in-dataset}, and \textsc{cross-data}, and we have the following findings:

1)	For \textsc{in-dataset} setting, comparing the \textit{Tag} and the basic models, we find a very simple method that assign each sample a domain tag could achieve improvement. The reason here we claim is that domain-aware model makes full use of the nature of dataset.
2)	For multi-domain and meta-learning model, we attempt to explain from the perspective of data distribution. Although meta-learning obtains worse performance under \textsc{in-dataset} setting, it yet has achieved impressive performance under \textsc{cross-dataset} setting.
Concretely, meta-learning model surpasses \textit{Tag} model on three datasets: \texttt{DUC2002}, \texttt{NYT50} and \texttt{Newsroom}, whose distribution is closer to CNN/DM based on \textit{constituent factors} in Table \ref{tab:PCR & CCR & R2}. Correspondingly, \textit{Tag} model uses a randomly initialized embedding for zero-shot transfer, and we suspect that this perturbation unexpectedly generalizes well on some far-distributed datasets (\texttt{arXiv} and \texttt{PubMed}).
3)	BERT has shown its superior performance and nearly outperforms all competitors. However, the generalization ability of BERT is poor on \texttt{arXiv}, \texttt{PubMed} and \texttt{DUC2002} compared to the performance improvement in \textsc{in-dataset} setting. In contrast, BERT shows good generalization when tranferring to datasets with high density and compression (\texttt{NYT50} and \texttt{Newsroom}). As we have discussed in Sec. \ref{p-sf}, samples with high \textit{style factors} require model to capture salient patterns, which is exactly the improvement of introducing external knowledge from BERT.

\subsubsection{Exp-II: Searching for a Good Domain}
The second question we study is  \textbf{what makes a good domain}?
To answer this question, we define the concept of domain based not solely on the dataset, but divide the training set by directly utilizing the \textit{constituent factors}.
Specifically, we explore the following different settings:

1) \textbf{Random tag:} Each sample is assigned a random ``\texttt{pseudo-domains}'' tag.

2) \textbf{Domain:} Divide training samples according to the domain (CNN or DM) they belong to .

3) \textbf{P- and C-Value:} Each sentence is assigned a tag by its corresponding P-Value and C-value scores.

\renewcommand\arraystretch{1.1}
\begin{table}[t]
\center \footnotesize
\tabcolsep0.07in
\setlength{\tabcolsep}{1mm}{
\setlength{\tabcolsep}{2.5mm}{
\begin{tabular}{llll}
\toprule
{Models} &
\textbf{R-1} & \textbf{R-2} & \textbf{R-L}  \\
\midrule

Transformer                & 41.31   & 18.71   & 37.62 \\
\quad + random tag               & 41.19   & 18.52   & 37.57 \\
\quad + domain tag          & \textbf{41.41}    & 18.71   & \textbf{37.74} \\
\quad + P-Value tag              & 41.38   & 18.71   & 37.67 \\
\quad + C-Value tag              & 41.39   & 18.73   & 37.71 \\
\quad + P-Value \& C-Value tag   & \textbf{41.41}    & \textbf{18.74} & \textbf{37.74} \\

\midrule

\citet{DBLP:journals/corr/abs-1903-10318}        & 42.57   & 19.96   & 39.04 \\
BERT (our implementation)  & 42.59   & 19.92   & 38.94 \\
\quad + domain tag          & 42.72   & 19.91   & 39.05 \\
\quad + P-Value \& C-Value tag   & \textbf{42.77}    & \textbf{19.98} & \textbf{39.10} \\

\bottomrule
\end{tabular}}}
\caption{Results of experiments with tags on our base model and current state-of-the-art model. The usage of BERT here is as same as \citet{DBLP:journals/corr/abs-1903-10318}, which is to fine tune BERT on CNN/DM.}
\label{table:different-tags}
\end{table}

We experiment with tags on our base model and the current state-of-the-art model \citet{DBLP:journals/corr/abs-1903-10318}.
\citet{DBLP:journals/corr/abs-1903-10318} and the results are presented in Table \ref{table:different-tags}, we can obtain the following observations:

1) \textit{Random partitioning does not make sense and cannot lead to the improvement of performance}. Conversely, the partitions based on the \textit{constituent factors} have obtained the benefit.
2) This simple learning method that dividing the training set based on domain has shown considerable benefit, which can be complementary to the improvement brought by BERT.
3) The division based on the \textit{constituent factors} (P-value \& C-value) achieves the best result in the context of BERT, which implies that \textit{for the summarization task, mining the characteristics of the dataset itself plays an important role}.

\section{Conclusion}
In this paper, we conduct a data-dependent understanding of neural extractive summarization models, exploring how different factors of datasets influence these models and how to make full use of the nature of the dataset so as to design a more powerful model.
Experiments with in-depth analyses diagnose the weakness of existing models and provide guidelines for future research.

\bibliographystyle{acl_natbib}
\bibliography{nlp}

\end{document}